\documentclass[lettersize,journal]{IEEEtran}
\pdfoutput=1
\usepackage{amsmath,amsfonts}
\usepackage{algorithmic}
\usepackage{algorithm}
\usepackage{array}
\usepackage{textcomp}
\usepackage{stfloats}
\usepackage{url}
\usepackage{verbatim}
\usepackage{graphicx}
\usepackage{cite}
\usepackage{epstopdf}
\usepackage{subfigure} 
\usepackage{multirow}
\usepackage{booktabs}
\begin{document}

\title{S2-Net: Self-supervision Guided Feature Representation Learning for Cross-Modality Images }

\author{~\IEEEmembership{Shasha Mei}
\thanks{}
\thanks{}}

\markboth{Journal of \LaTeX\ Class Files,~Vol.~14, No.~8, August~2021}%
{Shell \MakeLowercase{\textit{et al.}}: A Sample Article Using IEEEtran.cls for IEEE Journals}

\maketitle

\begin{abstract}

Combining the respective advantages of cross-modality images can compensate for the lack of information in the single modality, which has attracted increasing attention of researchers into multi-modal image matching tasks. Meanwhile, due to the great appearance differences between cross-modality image pairs, it often fails to make the feature representations of correspondences as close as possible. In this letter, we design a cross-modality feature representation learning network, S2-Net, which is based on the recently successful detect-and-describe pipeline, originally proposed for visible images but adapted to work with cross-modality image pairs. To solve the consequent problem of optimization difficulties, we introduce self-supervised learning with a well-designed loss function to guide the training without discarding the original advantages. This novel strategy simulates image pairs in the same modality, which is also a useful guide for the training of cross-modality images. Notably, it does not require additional data but significantly improves the performance and is even workable for all methods of the detect-and-describe pipeline. Extensive experiments are conducted to evaluate the performance of the strategy we proposed, compared to both handcrafted and deep learning-based methods. Results show that our elegant formulation of combined optimization of supervised and self-supervised learning outperforms state-of-the-arts on RoadScene and RGB-NIR datasets.
\end{abstract}

\begin{IEEEkeywords}
Cross-spectral, detect-and-describe, image mat-ching, deep learning, self-supervise learning.
\end{IEEEkeywords}

\section{Introduction}

Establishing the local correspondences between two images, as a primary task, is the premise of various visual applications, including target recognition, visual navigation, image stitching, 3D reconstruction and visual localization\cite{ma2021image,ohata2020automatic}. In particular, local feature representation is a popular approach to correspondence estimation. Recently, increasing attention has been focused on the cross-modality image feature representation learning and matching, since images of different modalities have their own advantages and contain different perceptual information, so it is meaningful to fuse the superiority of cross-modality images \cite{ma2019infrared}. 

The conventional matching methods are based on the handcrafted local feature descriptors \cite{ma2022robust,lowe2004distinctive,rublee2011orb,li2019rift} to make the representation of two matched features as similar as possible and as discriminant as possible from that of unmatched ones. Over the recent years, the deep learning-based methods have achieved significant progress in general visual tasks, and have also been introduced into the field of image matching \cite{yi2016lift,ono2018lf}. The current approaches are mostly based on a two-stage pipeline that first completes the extraction of keypoints and then encodes the patches centered on the keypoints into descriptors, thus referred to as the detect-then-describe methods. In the field of cross-modality image matching, the detect-then-describe methods have been widely used with a manual detector to detect and an adapted deep learning network to perform description \cite{jiang2021review,zhang2019registration,aguilera2017cross}. 

Despite this apparent success, it is an inevitable disadvantage of this paradigm that the global spatial information is discarded during the description process, which happens to be essential for cross-modality images. In contrast to it, the detect-and-describe framework for visible images uses a network to simultaneously perform feature point extraction and descriptor construction \cite{dusmanu2019d2,revaud2019r2d2}. This approach postpones the detection process without missing high-level information of images. Additionally, the detection stage is tightly coupled with the description so as to detect pixels with locally unique descriptors that are better for matching. Undoubtedly, it is promising to introduce the  framework into cross-modality image matching, however, challenges come up due to the huge heterogeneity. To be specific, it is difficult to optimize the model for cross-modality images with extreme geometric and radiometric variances. 

Self-supervised learning (SSL), which helps the model obtain easy invariance with augmented  data, is one of the most popular technics in natural language processing and computer vision\cite{chen2020simple,sermanet2017time}. As for local feature representation learning, the well-known Superpoint \cite{detone2018superpoint} proposed a novel Homographic Adaptation procedure, which is a form of self-supervision, to tackle the ill-posed problem of keypoint extraction. Nevertheless, the SSL technics have not been introduced into cross-modality scenario, while current methods are devoted to obtaining supervised signals from labeled data instead. Since the learning becomes harder for cross-modality images due to the serious radiometric variances, it is desirable to introduce SSL into this task. In fact, among the challenges faced by cross-modality descriptors, excluding inter-modal invariance, other necessities including geometric invariance as well as robustness to noise and grayscale variations can be well-addressed by SSL. The excellent data utilization efficiency and generalization ability of SSL enables its low-cost access to large amounts of training data, which is desperately required for cross-modality feature representation learning tasks.


\begin{figure}[!t]
	
	\centering
	\subfigure[]{
		\label{fig:a}
		\includegraphics[width=0.48\textwidth]{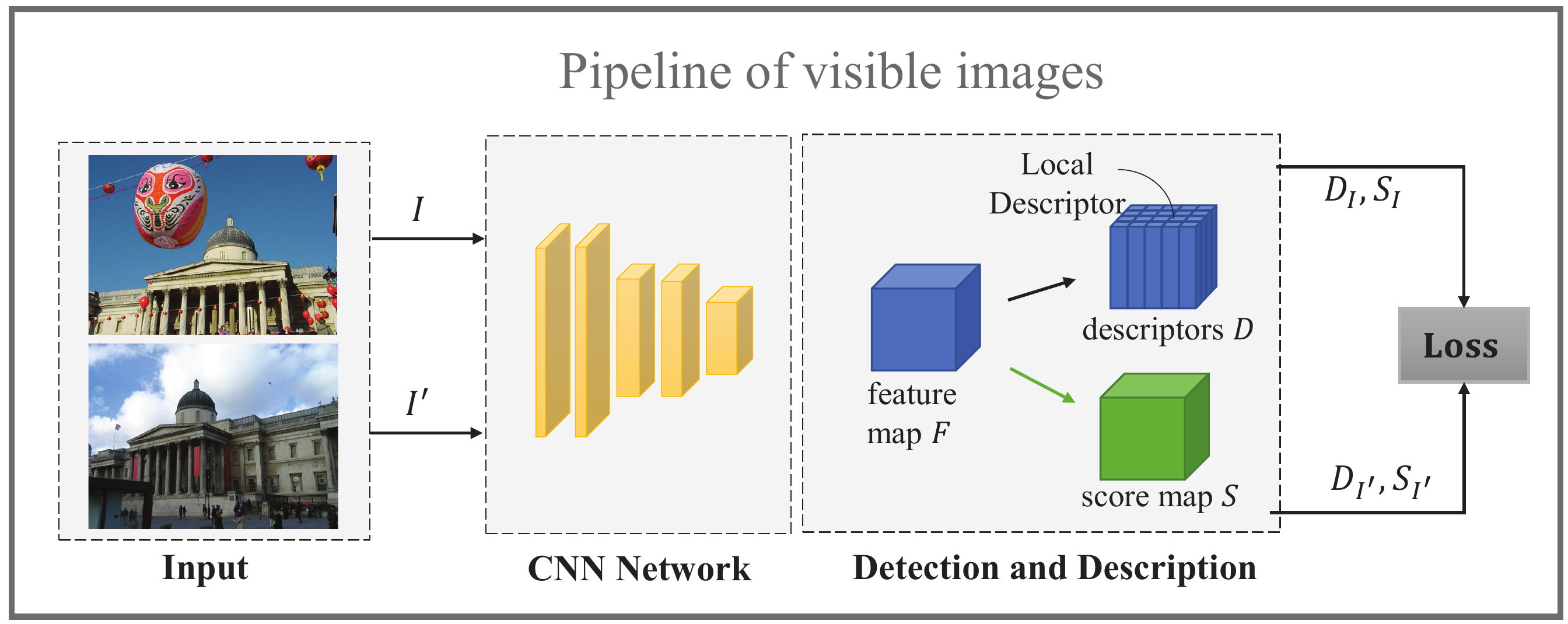} 
	}
	\subfigure[]{
		\label{fig:b}
		\includegraphics[width=0.48\textwidth]{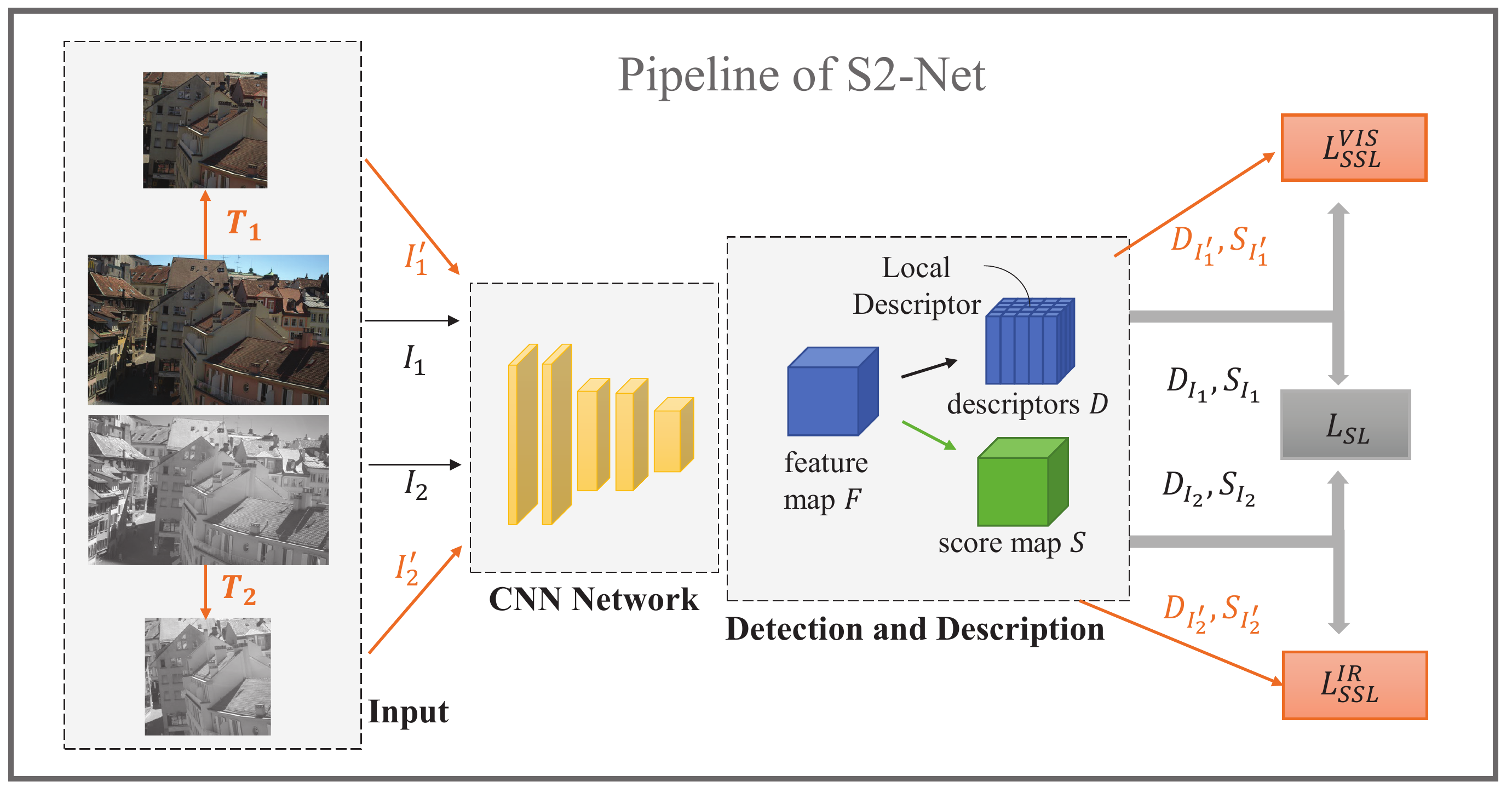}
	}
	\caption{Comparision of S2-Net and the classical pipeline of the detect-and-describe methods. (a) The pipeline of the detect-and-describe methods for visible images. (b)  Our proposed S2-Net for cross-modality images.} 
	\label{fig:1}
\end{figure}

In this work, we explore the possibility of using SSL, based on the recent success of the detect-and-describe methods, but adapted to work with cross-modality image pairs. Although the cross-modality images are heterogeneous and quite different in appearance, they still have some similar semantic information, such as the shape, structure, and topological relationship. The detect-and-describe methods retains the global spatial information which is rather crucial for our task. As for the optimization problem, we provide an effective solution for the application of the detect-and-describe methods to the cross-modality domain. More precisely, we propose our novel architecture of joint training with supervised and self-supervised learning, termed S2-Net, which takes full advantage of SSL to improve matching performance without extra labeled data, as illustrated in Fig. \ref{fig:1}. Self-supervision simulates the feature representation learning of images in the same modalities. Since the task of training image representations of the same modality is relatively easier compared to different modalities, self-supervision plays a guiding role in the training process. So in addition to supervised learning (SL), we add SSL to the optimization process, simultaneously training representations of images in the same and different modalities. Also, we design a loss function that combines both supervised and self-supervised learning and optimally balances the guidance of the two optimization methods. To the best of our knowledge, S2-Net is the first algorithm that introduces the SSL technique into cross-modality feature representation, and sufficient experiments have demonstrated the great effectiveness of our work.

\section{Method}
In this section, our proposed technique of self-supervision-guided optimization will be explained in detail, along with a delicated loss function.

\subsection{Pipeline of the detect-and-describe methods for visible images.}

Our proposed S2-Net , a general framework aims to make the detect-and-describe methods suitable for cross-modality image matching. The framework makes use of SSL to guide the training. Fig. \ref{fig:a} shows the classical pipeline of the detect-and-describe methods. The network takes a pair of corresponding images $(I, I^{'})$ and a correspondence $C$ between them as input and outputs their respective feature map $(F_{I}, F_{I^{'}})$. After that, the convolution operations with small number of parameters or non-maximum suppression are added to obtain sparse yet repeatable keypoint locations. As for description, nomalization is operated on the feature map as well as other optional constraints to choose descriptors. Finally, the computed descriptors $(D_{I}, D_{I^{'}})$ and score maps $(S_{I}, S_{I^{'}})$ form a loss function to penalize small distances between non-matching pairs, and large distances between matching pairs. It utilizes both local and global information and the network predicts a feature map both for detection and description so that the two processes can boost each other. 

\subsection{Framework of Self-supervision Guided Optimization}

To train the basic framework, relevant constraints for single modality images are proposed. However, the lack of strong supervision in these constraints, e.g., which point should be the key point, always troubles the training. Moreover, it is common knowledge that the difference between image pairs in the same modality is much smaller than that of cross-modality pairs. That is, it is somewhat difficult to train a good network for cross-modality feature representation, let alone a network that performs both detection and description. 
To this end, it is promising to introduce the mono-modality self-supervised learning to guide the cross-modality training. As illustrated in Fig. \ref{fig:b}, based on the basic framework in Fig. \ref{fig:a}, the other two branches with augmented cross-modality images for self-supervised learning are introduced for joint training, which is the major difference between S2-Net and the other methods.

Suppose we take the image pair $I$ as an example, and it is randomly tranformed to image $I'$ with the tranform $T(\cdot)$. Speficically, we focus on two major transforms, geometric transform and gray transform, which are the most natural discrepancies across modalities. As for geometric transforms, we mainly consider the random scale, random rotation and quadrangular random projection, which are denoted as $T_s(\cdot), T_r(\cdot)$ and $T_q(\cdot)$, respectively. Moreover, to encourage the model to obtain better generalization on the gray variation, we also add a random noise $T_n(\cdot)$, where the noise follows the normal distribution with the mean of 0 and the variance of 0.01, and a Gaussian blur $T_b(\cdot)$ with the scale of 1 as well as a random gray inverting $T_i(\cdot)$ that inverts the gray scale larger than a random threshold. Therefore, the total transform can be described in the cascading sub-transforms as: 
\begin{equation}
	\label{T3}
	T(\cdot)=T_s\cdot T_r\cdot T_q\cdot T_n\cdot T_b\cdot T_i(\cdot). 
\end{equation}
Independent random transform $T_{vis}$ and $T_{ir}$ would be conducted on raw visible and the other modality images for two parallel self-supervised learning.

%

%
By the above transformations, we artificially generate matching pairs that are close to the real existence, so as to guide the training of cross-modality image pairs. On the one hand, this approach retains the most important advantages of such one-stage methods, and on the other hand, it makes up for the difficulty of optimization.

\subsection{Overall Loss Function}
The input is processed by the network to generate the feature map, and then the descriptors and score maps are calculated. The addition of self-supervision can help guiding the optimization during training for cross-modality images. The overall loss function is expressed as follows:

\begin{equation}
	\label{loss}
	L = L_{SL} + \lambda (L_{SSL}^{VIS} + L_{SSL}^{IR}),
\end{equation}
where $L_{SL}$ is the SL loss, $L_{SSL}^{VIS}$ is the SSL loss for visible images, and $L_{SSL}^{IR}$ is the SSL loss for infrared images, thermal infrared or near-infrared in our experiments. $\lambda$ is the weighting coefficient of the loss functions, whose influence on the matching results is experimentally shown in the analysis of the weighting coefficients.

\subsubsection{SL loss}
Supervised learning is used in the original detect-and-describe methods, and also retained in our solution. Given a corresponding image pair $I_{vis}$ and $I_{ir}$, and $U$ denotes their mapping function. That is, for a pixel $p=(x_{i},y_{i})$ in $I_{vis}$ , $U_{(x_{i},y_{i})}$ is the corresponding pixel of $p$ in $I_{ir}$ . Then SL loss can be expressed as follows:
\begin{equation}
	\label{slloss}
	L_{SL} = L_{f}(I_{vis},I_{ir},U),
\end{equation}
where $L_{f}$ denotes the original loss function of a network $f(\cdot)$ in detect-and-describe methods for visible images. It is different from each method, but the inputs are fixed as a pair of corresponding images and the ground-truth correspondences between them. As a part of the overall loss, SL loss utilizes the labeled dataset to optimize the network. 

\subsubsection{SSL loss}
SL loss is common for detect-and-describe methods on visible images, however, in the cross-modality field, obtaining a large number of images with known correspondences requires a significant cost. In this case, SSL is a reliable and efficient solution, and we proposed the novel SSL loss: 
\begin{equation}
	\label{sllloss}
	\begin{split}
		L_{SSL}^{VIS} = L_{f}(I_{vis},I_{vis}^{'},T_{vis}),\\
		L_{SSL}^{IR} = L_{f}(I_{ir},I_{ir}^{'},T_{ir}),
	\end{split}
\end{equation}
where
\begin{equation}
	\label{T}
	\begin{split}
		I_{vis}^{'}=T_{vis}(I_{vis}),\\
		I_{ir}^{'}=T_{ir}(I_{ir}).
	\end{split}
\end{equation}
For visble images, $I_{vis}^{'}$ is obtained from $I_{vis}$ by transformation $T_{vis}$, so as to form a corresponding pair with $I_{vis}$ as the input to the network $f(\cdot)$. Also, the same operation is performed on the infrared images. It is worth mentioning that self-supervised learning does not rely on labeled data to work. It is useful not only to construct a bunch of artificial labels, but also to introduce homogeneous guidance in cross-modality learning that can guide the network closer to the correct representation.

\section{Experiments}
In order to demonstrate the effectiveness of our proposed approach, we selected D2-Net \cite{dusmanu2019d2} and R2D2\cite{revaud2019r2d2}, two classic detect-and-describe methods for visible images to compare the performance of our self-supervision on these methods. Furthermore, the handcrafted descriptors of SIFT \cite{lowe2004distinctive} and RIFT\cite{li2019rift} are added to the comparison experiments. To better evaluate the performance of our method on cross-modality images, we also compared CMM-Net \cite{cui2021cross}, which is a cross-modality feature representation network for thermal infrared and visible images.

\subsection{Implementation Details}
All the experiments were implemented on a computer with NVIDIA RTX 3090 GPU. 
The training process is based on the proposed loss, and the training setting varies with each achitecture. As for D2-Net, we have assessed the choice of the network used for feature extraction and then chosen VGG16 with the fully connection layers removed as the backbone architecture (following the settings in D2-Net). In particular, we fine-tuned the overall network for 100 epochs instead of fine-tuning the last layer of the dense feature extractor (conv4\_3). The network was optimized using Adam with a fixed learning rate of $10^{-4}$ and weight decay of $10^{-5}$. For each pair, we selected a random $256\times 256$ crop centered around one correspondence with a batch size of 1. In R2D2, the learning rate is 0.0001, the weight decay is 0.0005 and the baze size is 2 with the input pairs cropped to $192 \times 192$. The data augmentation is performed through the random flipping, random rotating $(90^{\circ}, 180^{\circ}, 270^{\circ})$ and random noise blurring. Moreover, the image pairs of the two datasets are co-registered with pixels aligned without offsets, so we additionally carried out random rotating $[-10^{\circ},10^{\circ}]$ , random scaling $[1,1.2]$ and random projection of a ratio $[0,0.2]$.

\subsection{Experimental Datasets}

\subsubsection{VIS-TIR Datasets}
The matching of thermal infrared (TIR) and visible images is a typical cross-modality problem, so we perform our experiments on RoadScene dataset\cite{xu2020fusiondn}, which is comprised of aligned thermal infrared and visible images. These images in this dataset are highly representative scenes from the FLIR video, with 221 aligned VIS and TIR image pairs containing rich scenes such as roads, vehicles, pedestrians and so on. The background thermal noise in the original TIR images has been pre-processed to accurately align the image pairs. Then the exact registration regions were cut out to form this dataset. This dataset was split in a testing dataset with 43 image pairs from different scenes and a training dataset from the remaining 178 pairs.

\subsubsection{RGB-NIR Datasets}
We also perform our experiments on a public registered RGB-NIR scene dataset by Aguilera \cite{brown2011multi}, which consists of 477 images in 9 scenes captured in RGB and Near-infrared (NIR). Since the RGB-NIR dataset covers different scenes of areas under different imaging codintions, it is suitable for our evaluation of robustness. We randomly select 171 images for testing (19 per scene) and train on the rest.

\begin{figure}[!t]
	
	\centering
	\subfigure[Results on VIS-TIR dataset when $K=4096$.]{
		\label{fig2_1}
		\includegraphics[width=0.5\textwidth]{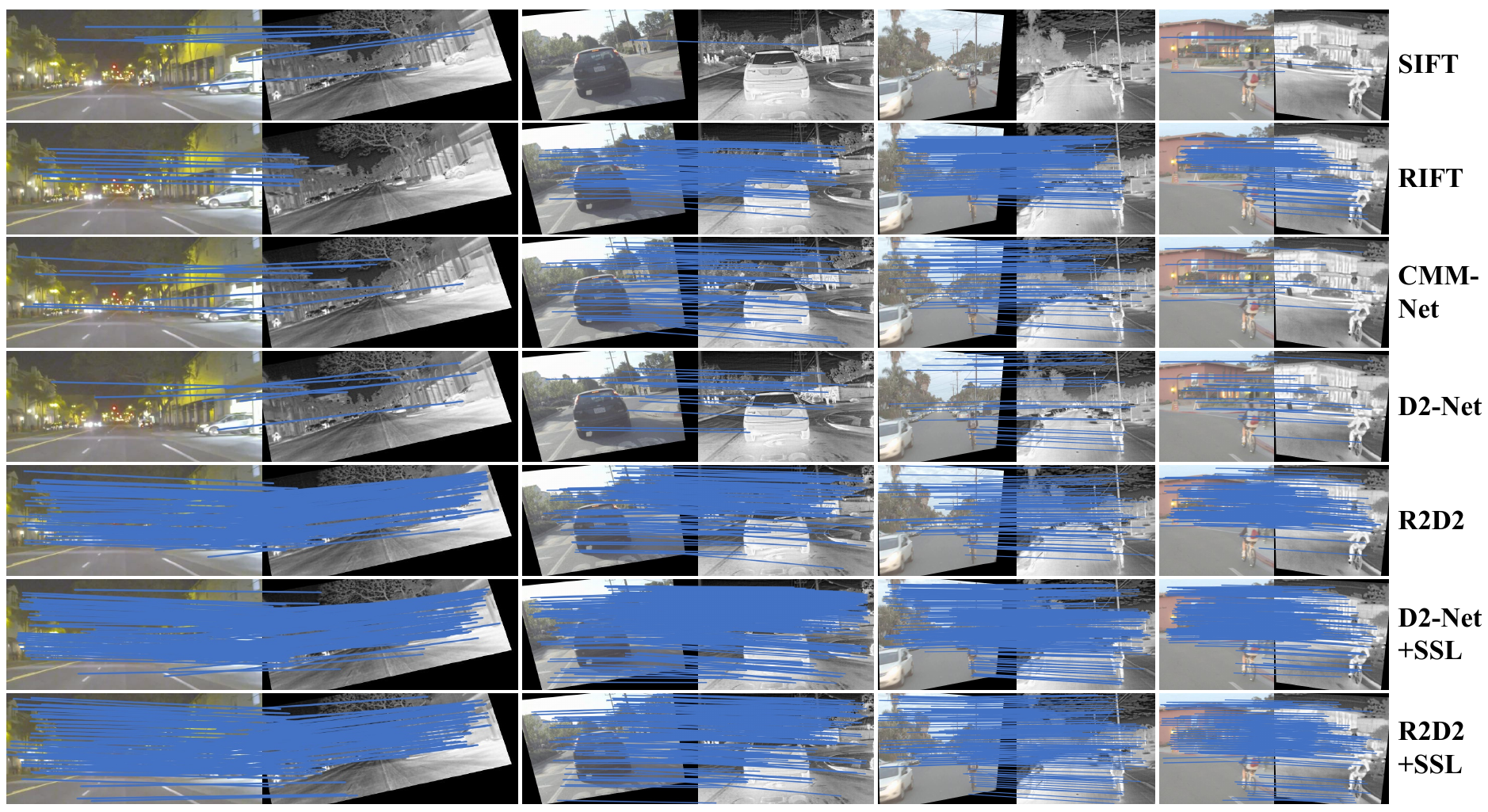}
	}%
	\quad
	\subfigure[Results on RGB-NIR dataset when $K=1024$.]{
		\label{fig2_2}
		\includegraphics[width=0.5\textwidth]{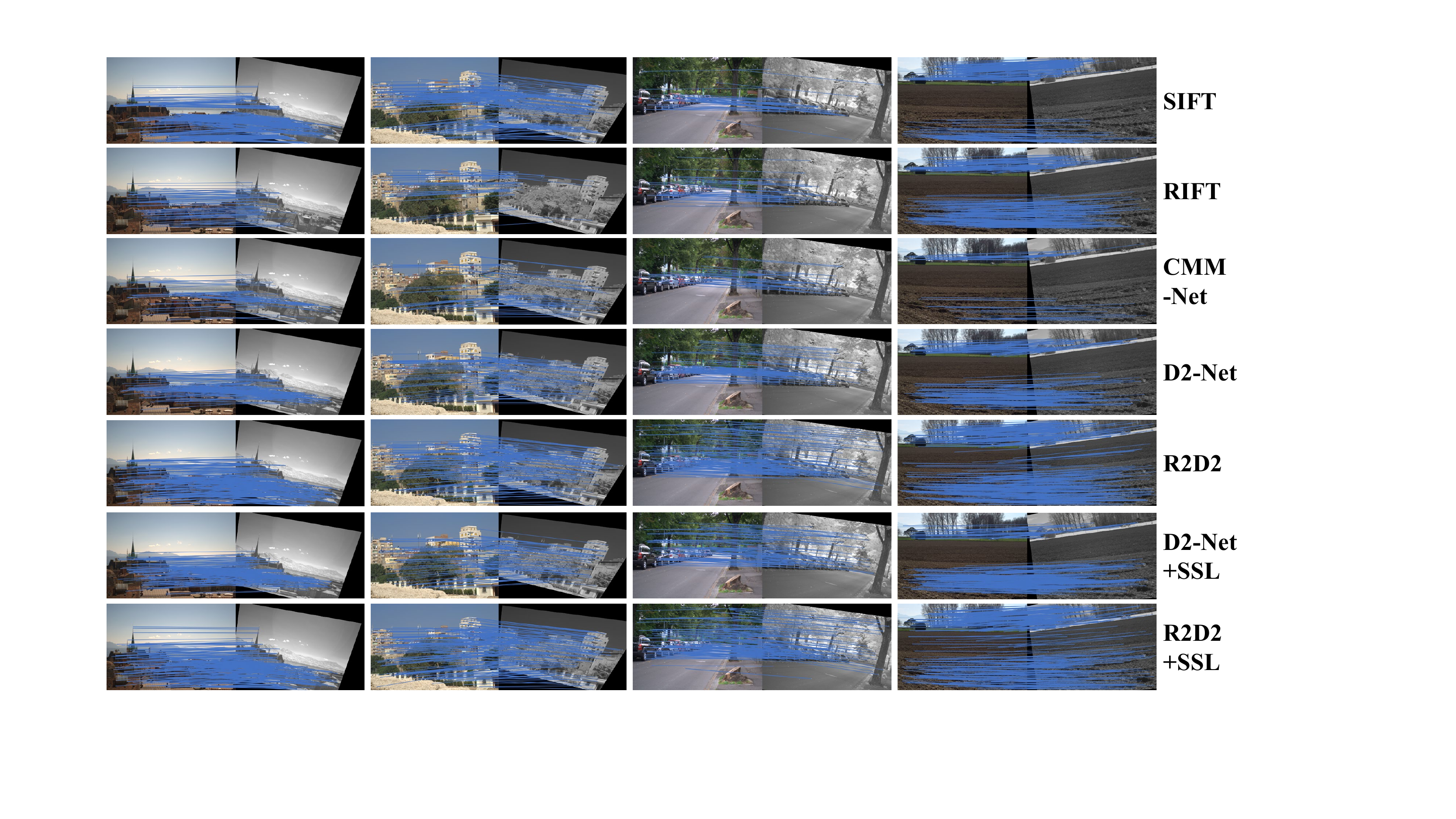}
	}
	\caption{Experimental results of S2-Net and the state-of-the-art image matching methods for the two datasets.} 
	\label{fig2}
\end{figure}

\subsubsection{Evaluation Metrics}
Three evaluation metrics, number of correspondences in the extracted points (NC), number of correct matches (NCM) and the correctly matched ratio (CMR) are used to evaluate the different methods quantitatively. NC indicates the repeatability of extracted interest points and NCM is crucial for the image registration. CMR is computed as
\begin{equation}
	\label{RCM}
	CMR=\frac{NCM}{NC}\times 100\%.
\end{equation}

\subsection{Comparison Results}
In the testing process, we vary the number of points extracted from both images, which is denonoted as K, and record the evalution results on each method. Specifically, we set $K=1024$, $K=2048$ and $K=4096$. The results obtained are listed in Table \ref{table1_1}-\ref{table1_3}. 
Combining the three evaluation metrics, it can be seen that on RoadScene dataset, D2-Net with the guide of SSL achieves the best performance, and R2D2 with SSL ranks second. After that, RIFT and CMM-Net, which were designed for cross-modality images, follow behind. The SIFT algorithm, performs the worst of all since it requires texture details that differ across modalities. It should be specially noted that the original R2D2 achieves fairly good results among the compared methods, but nevertheless, we improve it quite a bit. This is due to the fact that the original R2D2 algorithm takes repeatablity and reliability into account in its loss, which is not available in D2-Net. So with SSL, the performance of D2-Net has been extremely boosted. The relevant visualization results are shown in Fig. \ref{fig2_1}.

And on RGB-NIR dataset, since the difference between visible and thermal infrared images is much more significant than that with near-infrared images, SSL does not improve the performance of the original method as much as on the RoadScene dataset. And it is reasonable that SIFT achieves a good accuracy. However, the performance of R2D2 with SSL ranks best above all methods, as depicted in Fig. \ref{fig2_2}. The guiding effect of self-supervision in the training process is rather beneficial when learning modality-invariant feature representations.

\begin{table}[!t]
	\caption{Number of Correspondences in the extracted points on the two datasets.
		\label{table1_1}}
	\centering
	\resizebox{0.49\textwidth}{!}{
		\renewcommand\arraystretch{1.1}{
			
			\begin{tabular}{c|c|c|c||c|c|c}
				\toprule[1pt]
				\multirow{2}*{\textbf{Method}} & \multicolumn{3}{c||}{RoadScene dataset} & \multicolumn{3}{c}{RGB-NIR dataset}\\
				\cline{2-7}
				& $K=1024$ & $K=2048$ & $K=4096$ & $K=1024$ & $K=2048$ & $K=4096$ \\
				\midrule[0.9pt]
				SIFT & 294 & 788 & 1537 & 322 & 743 & 1681 \\
				RIFT & 443 & \textbf{1046} & 1295 & 351 & 845 & 1575 \\
				CMM-Net & 176 & 195 & 195 & 155 & 438 & 1146 \\
				D2-Net & 183 & 265 & 265 & 170 & 475 & 1283 \\
				R2D2 & 184 & 511 & 1669 & 448 & 953 & 2015 \\
				\midrule[0.9pt]
				\textbf{D2-Net+SSL} & \textbf{449} & 1033 & 1538 & 223 & 557 & 1424 \\
				\textbf{(ours)} & ($\uparrow$ 145.4\%) & ($\uparrow$ 289.8\%) & ($\uparrow$ 480.4\%) & ($\uparrow$ 31.2\%) & ($\uparrow$ 17.3\%) & ($\uparrow$ 11.0\%) \\
				\hline
				\textbf{R2D2+SSL} & 243 & 668 & \textbf{2015} & \textbf{474} & \textbf{1013} & \textbf{2165} \\
				\textbf{(ours)} & ($\uparrow$ 32.1\%) & ($\uparrow$ 30.7\%) & ($\uparrow$ 20.7\%) & ($\uparrow$ 5.8\%) & ($\uparrow$ 6.3\%) & ($\uparrow$ 7.4\%) \\
				\bottomrule[1pt]		
				
	\end{tabular}}}
\end{table}

\begin{table}[!t]
	\caption{Number of points Correctly Matched on the two datasets.
		\label{table1_2}}
	\centering
	\resizebox{0.49\textwidth}{!}{
		\renewcommand\arraystretch{1.1}{
			\begin{tabular}{c|c|c|c||c|c|c}
				\toprule[1pt]
				\multirow{2}*{\textbf{Method}} & \multicolumn{3}{c||}{RoadScene dataset} & \multicolumn{3}{c}{RGB-NIR dataset}\\
				\cline{2-7}
				& $K=1024$ & $K=2048$ & $K=4096$ & $K=1024$ & $K=2048$ & $K=4096$ \\
				\midrule[0.9pt]
				SIFT & 3 & 6 & 6 & 141 & 286 & 541 \\
				RIFT & 36 & 57 & 66 & 111 & 200 & 298 \\
				CMM-Net & 29 & 31 & 31 & 36 & 95 & 233 \\
				D2-Net & 12 & 14 & 14 & 90 & 214 & 508 \\
				R2D2 & 30 & 60 & 157 & 234 & 458 & 825 \\
				\midrule[0.9pt]
				\textbf{D2-Net+SSL} & \textbf{92} & \textbf{177} & \textbf{233} & 127 & 272 & 590 \\
				\textbf{(ours)}& ($\uparrow$ 666.7\%) & ($\uparrow$ 1157.1\%) & ($\uparrow$ 1564.3\%) & ($\uparrow$ 41.1\%) & ($\uparrow$ 27.1\%) & ($\uparrow$ 16.1\%) \\
				\hline
				\textbf{R2D2+SSL} & 50 & 93 & 228 & \textbf{248} & \textbf{486} & \textbf{903} \\
				\textbf{(ours)}& ($\uparrow$ 66.7\%) & ($\uparrow$ 55.0\%) & ($\uparrow$ 45.2\%) & ($\uparrow$ 6.0\%) & ($\uparrow$ 6.1\%) & ($\uparrow$ 9.5\%) \\
				\bottomrule[1pt]		
				
	\end{tabular}}}
\end{table}

\begin{table}[!t]
	\caption{Ratio of Correct Matches on the two datasets.
		\label{table1_3}}
	\centering
	\resizebox{0.49\textwidth}{!}{
		\renewcommand\arraystretch{1.1}{
			\begin{tabular}{c|c|c|c||c|c|c}
				\toprule[1pt]
				\multirow{2}*{\textbf{Method}} & \multicolumn{3}{c||}{RoadScene dataset} & \multicolumn{3}{c}{RGB-NIR dataset}\\
				\cline{2-7}
				& $K=1024$ & $K=2048$ & $K=4096$ & $K=1024$ & $K=2048$ & $K=4096$ \\
				\midrule[0.9pt]
				SIFT & 1.37 & 0.76 & 0.41 & 36.97 & 33.45 & 28.78 \\
				RIFT & 8.11 & 4.98 & 4.49 & 28.17 & 21.61 & 17.46 \\
				CMM-Net & 16.66 & 15.62 & \textbf{15.62} & 23.58 & 21.68 & 20.16 \\
				D2-Net & 6.37 & 5.25 & 5.25 & 51.00 & 23.22 & 31.33 \\
				R2D2 & 16.03 & 11.89 & 9.42 & 47.04 & 42.77 & 36.27 \\
				\midrule[0.9pt]
				\textbf{D2-Net+SSL} & \textbf{20.50} & \textbf{17.04} & 14.46 & \textbf{56.76} & \textbf{48.05} & \textbf{40.71} \\
				\textbf{(ours)}& ($\uparrow$ 221.8\%) & ($\uparrow$ 224.6\%) & ($\uparrow$ 175.4\%) & ($\uparrow$ 11.3\%) & ($\uparrow$ 106.9\%) & ($\uparrow$ 29.9\%) \\
				\hline
				\textbf{R2D2+SSL} & 20.17 & 13.95 & 11.32 & 48.94 & 44.97 & 38.94 \\
				\textbf{(ours)}& ($\uparrow$ 25.8\%) & ($\uparrow$ 17.3\%) & ($\uparrow$ 20.2\%) & ($\uparrow$ 4.4\%) & ($\uparrow$ 5.1\%) & ($\uparrow$ 7.4\%) \\
				\bottomrule[1pt]		
				
	\end{tabular}}}
\end{table}


\section{Discussion} \label{value}
\begin{figure}[h]
	\centering
	\subfigure[]{
		\includegraphics[width=0.24\textwidth]{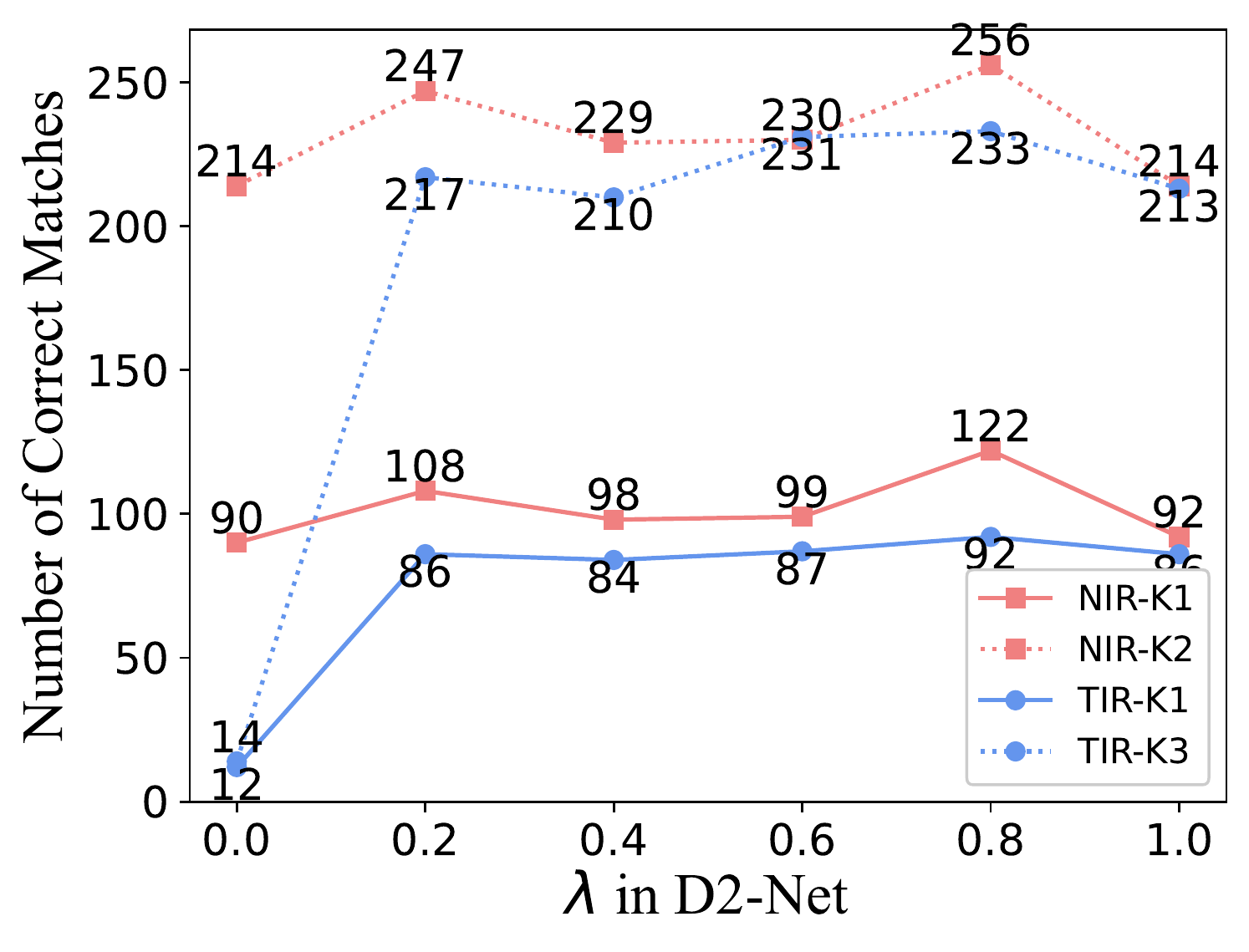}
	}%
	\subfigure[]{
		\includegraphics[width=0.24\textwidth]{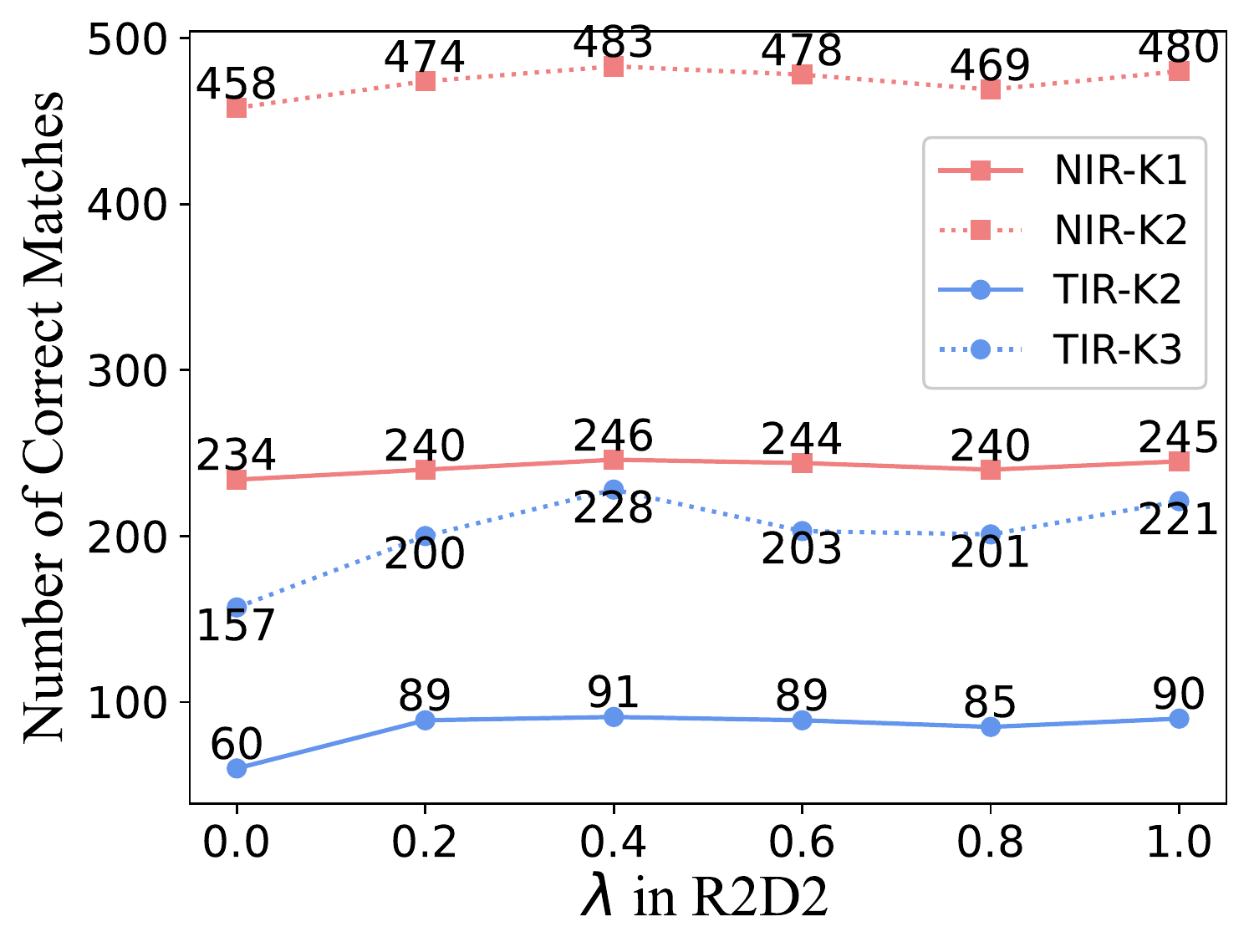}
	}
	\caption{ Evaluation of the weighting coefficients $\lambda$ in D2-Net and R2D2 on the RoadScene dataset and RGB-NIR dataset. }
	\label{fig3}
\end{figure}

Since we carry out the network learning with joint training of supervised and self-supervised learning, it is worth to evaluate the key weighting coefficient $\lambda $ of the SSL loss in the overall loss function. Increasing the parameter is to magnify the guiding effect of self-supervision on the training process, and vice versa. We vary the value $\lambda$ and record the number of correct matches(NCM), which has been depicted in Fig. \ref{fig3}. To be more comprehensive, we set the extracted points $K=1024,2048$ and 4096, corresponding to K1, K2 and K3 in Fig. \ref{fig3}, respectively, and then select some displayed for the best view. 

The impact of self-supervision varies with different methods at different settings. Since D2-Net completes the network learning with a score-weighted triplet margin ranking loss, it ignores the repeatability constraints, which requires compensations from self-supervised learning. As a result, when $\lambda=0.8$, it achieves the best results on D2-Net. In contrast, SSL on R2D2 offers optimal performances when $\lambda=0.4$, indicating that the effect of self-supervised learning is similar to that of supervised learning. R2D2 jointly learns the estimations of the keypoint repeatability and the descriptor reliability, so that it is able to lead to the right goal of network learning, and therefore receives little influence from the weighting coefficient $\lambda$.

%

\section{Conclusion}
In this article, we proposed a joint learning strategy of supervised learning and self-supervised learning (S2-Net) based on the detect-and-describe methods to learn the modality-invariant feature representation. We employ a novel loss which introduces the self-supervised learning in the training to guide the network learning. It is a simulation of mono-modal images and encourages feature representation learning in the same and different modalities. After performing experiments on RoadScene and RGB-NIR datasets, it can be demonstrated that our strategy significantly improves the networks' capability of feature representation for cross-modality images, including the detection and description. Notably, although the proposed method has shown promising results, our work is based on the detect-and-describe methods, so an obvious direction for the future work is to explore more possible forms that can be transfered to more methods.

\bibliographystyle{IEEEtran}
\bibliography{ref}
\end{document}